\documentclass[letterpaper]{article} 
\usepackage[preprint]{aaai2027}  
\usepackage[hyphens]{url}  
\usepackage{graphicx} 
\usepackage{natbib}  
\usepackage{caption} 
\usepackage{algorithm}
\usepackage{algorithmic}
\usepackage{array} 
\usepackage[table]{xcolor}
\definecolor{rowblue}{RGB}{220,230,241} 
\usepackage{makecell}
\usepackage{amsmath}
\usepackage{multirow}
\usepackage{pifont}
\newcommand{\name}{\emph{OmniRouting}}

\usepackage{subcaption}
\usepackage{newfloat}
\usepackage{listings}
\DeclareCaptionStyle{ruled}{labelfont=normalfont,labelsep=colon,strut=off} 
\floatstyle{ruled}
\newfloat{listing}{tb}{lst}{}
\floatname{listing}{Listing}

\usepackage{booktabs}

\title{OmniRouting: A Semantic-Coupled Multimodal Benchmark for Constraint-Aware Spatial Reasoning in PCB Routing}

\author{
    Taiting Lu\textsuperscript{\rm 1}\equalcontrib,
    Kaiyuan Lin\textsuperscript{\rm 1}\equalcontrib,
    Ziwei Dong\textsuperscript{\rm 2},
    Sisong Bei\textsuperscript{\rm 2},
    Haolin Ye\textsuperscript{\rm 1},
    Yuxin Tian\textsuperscript{\rm 2},
    Runze Liu\textsuperscript{\rm 1},
    \\
    Mingjia Wang\textsuperscript{\rm 3},
    Jingying Zeng\textsuperscript{\rm 2},
    Hongxing Pan\textsuperscript{\rm 3},
    Kai Zhang\textsuperscript{\rm 3},
    Haoyu Wang\textsuperscript{\rm 1},
    Guoliang Shi\textsuperscript{\rm 3},
    \\
    Ling Ma\textsuperscript{\rm 3},
    Yifan Yang\textsuperscript{\rm 4},
    Jiaying Lu\textsuperscript{\rm 5},
    Qi He\textsuperscript{\rm 2},
    Yi-Chao Chen\textsuperscript{\rm 3},
    Sung-Liang Chen\textsuperscript{\rm 3},
    Yincheng Jin\textsuperscript{\rm 6},
    Mahanth Gowda\textsuperscript{\rm 1}\corresponding
}

\affiliations{
    \textsuperscript{\rm 1}Pennsylvania State University,
    \textsuperscript{\rm 2}Independent Researcher,
    \textsuperscript{\rm 3}Shanghai Jiao Tong University\\
    \textsuperscript{\rm 4}Microsoft Research,
    \textsuperscript{\rm 5}Emory University,
    \textsuperscript{\rm 6}Binghamton University
}

\begin{document}

\maketitle

\begin{abstract}
Recent large language models (LLMs) have demonstrated remarkable progress in constraint-aware navigation, maze reasoning, graph reasoning. 
However, their ability to reason about complex routing problems under strict geometric, topological, and electrical constraints remains largely unexplored, despite routing being one of the most challenging and critical stages of electronic design automation (EDA). 
To bridge this gap, we introduce \textbf{\name}, the first large scale benchmark designed to evaluate LLMs on printed-circuit-board (PCB) routing reasoning under real-world industrial design-rule, manufacturability, and connectivity constraints.
\textbf{\name} contains 1,681 industrial-grade schematic-coupled PCB designs, including board geometries, rout-able component placements by human engineer, footprints, pad locations, netlists, stackup information, and routing constraints. 
The benchmark comprises four tasks: \textbf{(1) geometric routing reasoning}, generating physically valid copper traces, vias, and layer assignments to connect circuit nets within constrained board regions; \textbf{(2) design-rule-aware routing reasoning}, producing routable layouts that satisfy clearance, trace-width, via, obstacle-avoidance, and board-boundary constraints; \textbf{(3) electrical functionality reasoning}, preserving schematic-specified connectivity while reasoning over net names and functional roles to produce electrically correct routing; and \textbf{(4) tool-augmented agentic routing}, leveraging external tools for (1)-(3).
Our results reveal substantial limitations of current LMMs in PCB routing, including weak path-planning capabilities, poor adherence to design-rule constraints, and inconsistent preservation of electrical functionality. We will open-source all benchmark data, evaluation code, and tool interfaces to facilitate future research.
\end{abstract}

\section{Introduction}\label{intro}

\begin{figure}[!t]
  \centering
  \includegraphics[width=0.9\columnwidth]{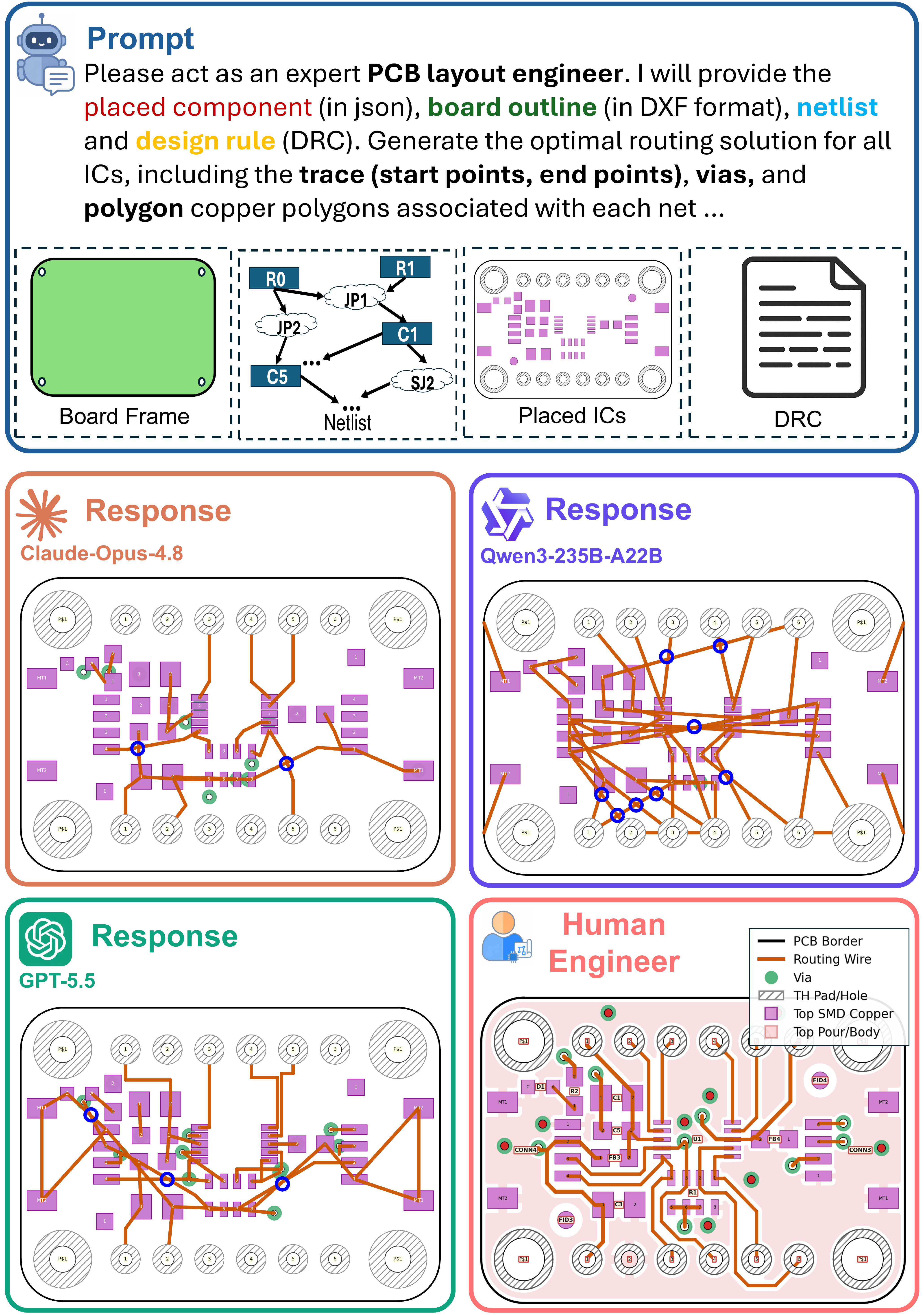}
  \caption{Large language models (LLMs) fail to perform reliable PCB routing under complex geometric, electrical, and manufacturing constraints. Frequent routing violations (highlighted by dashed \textcolor{blue}{blue} circles), including unrouted connections, trace conflicts, and constraint violations, reveal substantial weaknesses in spatial reasoning and constraint-aware routing generation.}
  
  \label{fig:lmm_test}
\end{figure}

As the scale, integration level, and component density of integrated circuits (ICs) continue to increase, the complexity of their physical realization on printed circuit boards (PCBs) grows accordingly. Since PCBs provide the physical interconnections required to realize system functionality, PCB EDA tools play a critical role in modern electronic design. Among the stages of the PCB design flow, routing is one of the most challenging and computationally intensive combinatorial optimization problems. Given a fixed component placement and connectivity specification, routing generates conductive interconnects between component pins while minimizing wirelength and satisfying manufacturing and design constraints.
Unlike IC routing, which benefits from highly standardized design abstractions and consequently achieves a high degree of automation, PCB routing remains heavily dependent on expert layout engineers because of its inherently irregular geometries, diverse component layouts, and complex routing constraints.
Additionally, while IC routing research benefits from extensive benchmark suites comprising hundreds or even thousands of designs \cite{alpert1998ispd98,nam2008ispd,mantik2018ispd}, publicly available PCB routing datasets remain scarce. Because industrial PCB layouts are almost exclusively proprietary, most existing PCB routing studies \cite{xie2018routenet,chen2020pros,li2023fanoutnet} rely on fewer than 20 real-world board designs for training and evaluation.

In recent years, large multimodal models (LMMs) \cite{chen2021topological,chen2024spatialvlm,yuan2025opennav,dao2025alphamaze}, have demonstrated strong capabilities in constraint-aware navigation, visual spatial reasoning, topological planning, and fine-grained geometry. 
These advances suggest new opportunities for applying LMMs to PCB routing, which can be viewed as a constrained path-planning problem requiring the generation of valid trace paths while satisfying diverse constraints.
However, existing spatial-LMM works largely overlook the core difficulty of PCB routing: generating valid trace paths in irregular, congested layouts while satisfying design-specific constraints such as clearance, trace width, layer transitions, differential pairs, length matching and signal integrity. 
These constraints are difficult to fully formalize and often require expert routing knowledge.
This raises a critical open question: \textbf{\textit{Can current LMMs generate PCB routing solutions that satisfy geometric constraints, preserve electrical connectivity, and remain functional across diverse real-world circuit designs?}}

To investigate this question, we conducted a series of preliminary experiments to assess the PCB routing performance of several state-of-the-art LMMs, including Claude-Opus-4.8 \cite{anthropic_claude_opus_47}, GPT-5.5 \cite{openai_gpt55}, and Qwen3.6-35B-A3B \cite{qwen3-35b-a3b}. Their performance was compared against that of experienced human layout engineers.
Specifically, we evaluated their ability to generate routing solutions for real-world PCB designs with engineer-crafted component placements that have been verified to be fully routable. This setup isolates the routing task from placement quality and ensures that any failures arise from routing rather than unroutable component arrangements.
As illustrated in Figure \ref{fig:lmm_test}, all evaluated models frequently produce routing failures, including unrouted connections, trace conflicts, and design-rule violations, such as routing GND as ordinary traces instead of implementing continuous ground polygons and return-current paths.
These results reveal that current LMMs fundamentally lack the topological planning and constraint-aware navigation capabilities required for PCB routing, frequently failing to generate complete, conflict-free, and electrically functional routing solutions in congested routing environments.

\begin{figure*}[!t]
    \centering
    \includegraphics[width=0.9\textwidth]{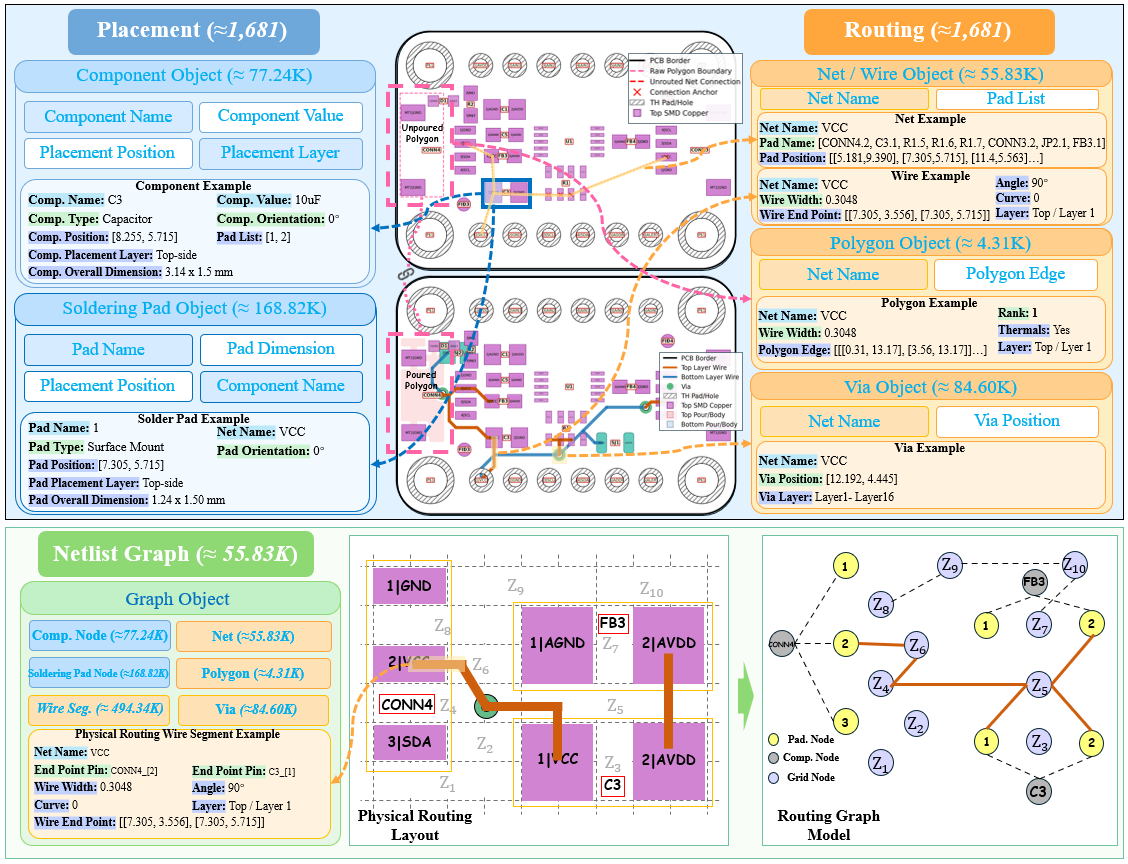}
    \caption{Overview of {\name} benchmark with representative cases.}
    \label{fig:dataset_overview}
\end{figure*}

To bridge this gap, we introduce \textbf{\name}, the first large-scale benchmark for evaluating LMMs on schematic-coupled PCB routing reasoning, comprising industrial-grade designs with ensured routable component layouts and complete human engineer reference routing implementations, including all connection geometries (trace), layer transitions (via), and conductive structures (polygon) required for physical realization.
As shown in Figure \ref{fig:dataset_overview}, {\name} contains 1,681 real-world PCB layouts paired with corresponding schematics and expert-designed reference layouts, spanning a wide range of board areas ($31.53--18,903.00\,mm^2$) and PCB stackups from 2 to 8 layers.
The benchmark comprises 77.24\textit{K} component placement instances, 168.82\textit{K} pads and SMDs, 55.83\textit{K} semantically annotated net connections, 84.60\textit{K} vias, and 4.31\textit{K} copper pours, capturing complete PCB routing implementations, including traces, vias, and plane geometries, that pass industrial design-rule checks (DRC) and satisfy fabrication constraints.
We first propose a systematic evaluation protocol for assessing PCB routing designs generated by LMMs with respect to connectivity, routing legality, and electrical functionality.
Our benchmark evaluates four core abilities, including
\textbf{(i) topological path planning} for constructing complete routing topologies that connect circuit nets,
\textbf{(ii) design-rule-aware routing reasoning} for generating manufacturable routing solutions under geometric and fabrication constraints,
\textbf{(iii) electrical functionality reasoning} preserving the electrical intent of expert-designed layouts, as measured by routing-path, via-placement, and topology similarity to reference designs;
and \textbf{(iv) tool-augmented agentic routing} for iterative routing refinement through benchmark-provided routing analysis and optimization tools.

Our main contributions are as follows:
\textbf{(i)} We introduce \textbf{\name}, a comprehensive benchmark for evaluating LMMs on PCB routing reasoning, containing 1,681 real-world schematic-coupled PCB designs with expert-designed reference routings, complete ground-truth traces, vias, and copper pours, as well as systematic evaluation protocols for routing completion, connectivity, design-rule compliance, routing quality, and reference similarity.
\textbf{(ii)} We systematically evaluate state-of-the-art LMMs on PCB routing under multiple settings, including zero-shot routing generation and tool-augmented agentic routing with iterative visualization, scoring, and routing-refinement feedback.
\textbf{(iii)} We provide a detailed analysis of LMM capabilities in PCB routing, examining performance across topological routing reasoning, design-rule-aware routing generation, preservation of electrical intent, routing quality optimization, and the effectiveness of tool-augmented agentic routing through iterative interaction with routing analysis tools.
\section{Related Work}\label{related_work}

\textbf{Topological Planning in LMMs.}
Recent LLMs have shown growing capabilities in constraint-aware navigation, maze solving, and graph reasoning. Early studies such as NavGPT and PPNL evaluate instruction following, feasible path planning, obstacle avoidance, and spatial-temporal constraint satisfaction in embodied or grid-world environments \cite{zhou2024navgpt,aghzal2023can}. Later work examines maze reasoning and spatial state tracking through visualization-based prompting and coordinate-based benchmarks \cite{wu2024mind,einarsson2026mazeeval}. In parallel, graph benchmarks including NLGraph, Talk like a Graph, GraphInstruct, GraphWiz, and GraphArena test connectivity, shortest paths, traversal, flow, and other combinatorial problems \cite{wang2023can,fatemi2024talk,luo2024graphinstruct,chen2024graphwiz,tang2025grapharena}.
Despite this progress, LMM research has largely overlooked PCB routing, which is substantially more complex than conventional constrained navigation. Instead of planning a single collision-free path, PCB routing requires jointly optimizing thousands of interdependent routes under heterogeneous geometric, electrical, and manufacturability constraints.

\textbf{Benchmarks for PCB Routing.} 
In recent years, commercial tools \cite{cadence2026allegroxai,topor2026,freerouting2026,quilter2026} and academic methods \cite{liao2020deep,he2020circuit,tseng2015ilp,fang2024obstacle} have increasingly automated PCB routing through maze routing, rip-up-and-reroute, push-and-shove, constraint-based optimization, and learning-based approaches. Commercial systems such as Cadence Allegro X AI, TopoR, FreeRouting/ELECTRA, and Quilter offer automatic or AI-assisted routing, but are typically developed within proprietary workflows whose board designs, constraints, layer stackups, design rules, and routing outputs are not publicly available. This limits reproducible benchmarking and fair comparison.
Academic work has explored reinforcement learning for global routing \cite{liao2020deep}, MCTS-guided neural routing \cite{he2020circuit}, ILP-based meander optimization \cite{tseng2015ilp}, and obstacle-aware length matching \cite{fang2024obstacle}. However, existing public benchmarks remain small and simplified, often containing no more than 20 mostly two-layer boards. PCB routing therefore still lacks an open benchmark of realistic multi-layer industrial designs with complex placements, heterogeneous net classes, and diverse clearance constraints.

\begin{figure}[t]
    \centering
    \includegraphics[width=\columnwidth]{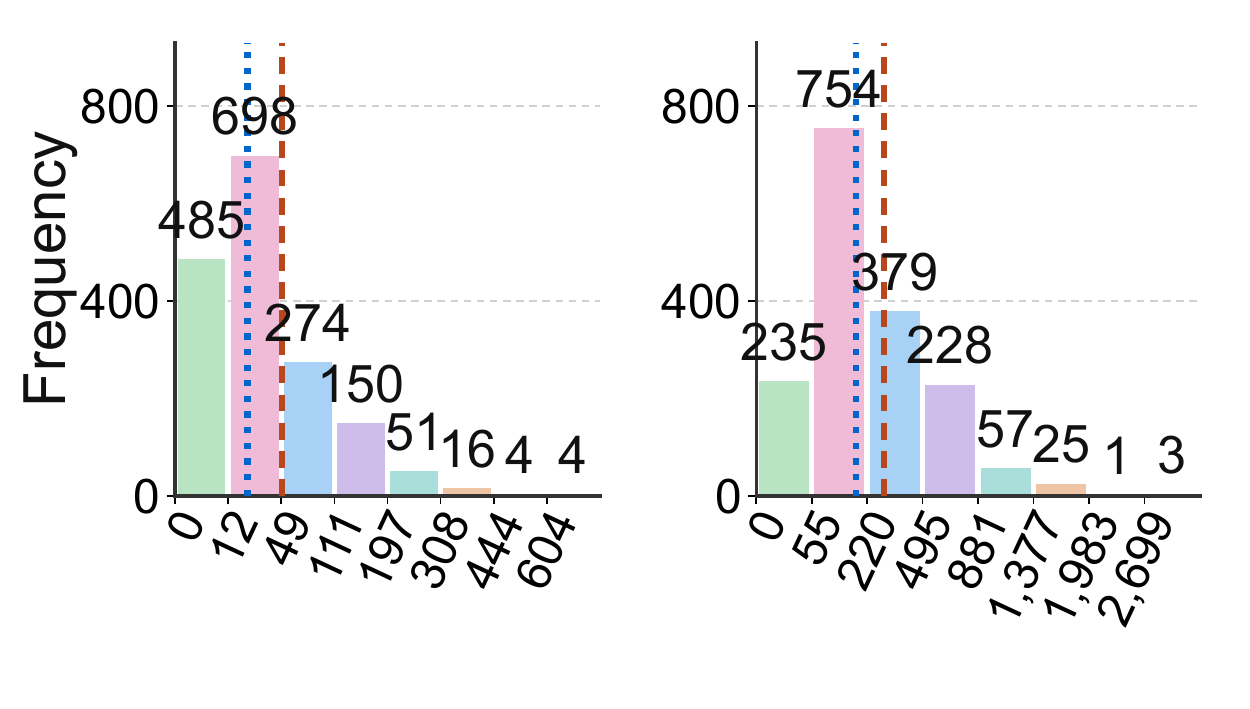}
    \caption{Routing statistics of the {\name} benchmark. Left: routing vias per PCB layout. Right: straight routing segments per PCB layout. Orange dashed lines indicate the mean; blue dotted lines indicate the median.}
    \label{fig:routing_statistics}
\end{figure}

\section{Benchmark Construction}\label{benchmark}

\textbf{Task Formulation.}
To provide a comprehensive evaluation framework for PCB routing tasks, our task formulation covers four key capabilities of constraint-aware geometric reasoning:
\textbf{(i)} \textbf{Connectivity}, which requires generating routing paths that completely connect every net while preserving the intended electrical topology, evaluated by routability, open count, and short count;
\textbf{(ii)} \textbf{Design-Rule Compliance}, which requires producing manufacturable routing that satisfies physical design constraints, evaluated by total DRC, clearance, trace-width, via-rule, and boundary/keepout violations;
\textbf{(iii) Electrical Functionality}, which assesses whether the routing correctly realizes the intended behavior of each net based on the circuit connectivity and semantic information (e.g., net names and pin functions).
\textbf{(iv)Agentic Tool Use for Constraint-Aware Geometric Reasoning}, which formulates PCB routing as an iterative decision-making problem, where LMMs are equipped with tools (such as routing visualization, conflict checking, and connectivity verification) to progressively refine routing paths until all geometric and electrical constraints are satisfied.

\textbf{Annotation Curation.}
We construct the {\name} benchmark using a scalable automated pipeline built on OmniSch~\cite{lu2026omnisch} and OmniLayout~\cite{lu2026omnilayout}. 
We collect 1,681 real-world schematic designs from open-source hardware platforms, including SparkFun, Arduino, Adafruit, GitHub, Seeed Studio, and ProtoCentral~\cite{sparkfun, github, arduino, adafruit, seeedstudio, protocentral}.
While OmniLayout only supports PCB layout rendering for component placement, we develop a fully functional PCB EDA engine supporting both component placement and routing, capable of parsing source-level design information directly from EAGLE XML files, including component placements, board geometry, routing wires, vias, and copper polygons. 
Beyond annotation extraction, the engine also serves as an interactive visualization tool in our agentic evaluation framework, rendering the current routing state as visual feedback for LMMs.

\textbf{Statistics of OmniRouting Benchmark.} 
As illustrated in Figure~\ref{fig:dataset_overview} and Figure~\ref{fig:routing_statistics}, {\name} contains 1,681 real-world schematic-coupled PCB designs covering a broad range of application domains, including robotics, wireless systems, and sensing, providing a comprehensive benchmark for supervised learning and evaluation of PCB routing, and related EDA tasks.
Our dataset includes three levels of structured annotations: 
\textbf{(i)} \textbf{ Comprehensive Routing Task}, providing fine-grained annotations for 77,242 components, 168,815 pads, and 55,830 nets, paired with aligned schematics and PCB layouts. Every PCB design is verified to be fully routable. Each component is annotated with its identity, value, placement, orientation, and layer assignment.
\textbf{(ii)} \textbf{Reference Routing Annotations}, including a engineered verified reference routing consisting of routing 494,349 wire segments, 84,603 vias, and 4,314 copper polygons, together with their geometric and manufacturing attributes (e.g., layer assignment, trace width, routing angle, via properties, and polygon boundaries). 
These reference solutions enable both functional evaluation and reference-based routing similarity evaluation.


\section{Evaluation Metrics}

Evaluating PCB routing requires jointly considering electrical connectivity, physical design rules, and routing efficiency. PCB routing is a constrained combinatorial optimization problem with NP-hard subproblems, and its search space grows rapidly with routing corridors, layer assignments, via locations, and trace geometries. Because multiple layouts may satisfy the same netlist and physical constraints, the human-engineered layout is treated as one valid expert reference rather than a unique geometric ground truth.
This geometric non-uniqueness makes exact trace matching unsuitable: small variations or alternative paths may remain fully valid. At the same time, connectivity and design-rule checks alone cannot capture meaningful spatial-routing decisions, such as choosing different corridors around an intermediate component. We therefore separate \emph{absolute routing validity} from \emph{reference-relative routing agreement}. The former evaluates correctness against the expected inputs, target netlist, and published design rules, while the latter measures similarity to the routing strategy demonstrated by the reference.

For absolute evaluation, Pass Rate (PR) measures load and schema success, including outputs with partial evaluation errors. Net Routability Ratio (NRR) is the percentage of expected nets that are connected and free of published DRC violations, while Pad-to-Pad Routability Ratio (PRR) measures the percentage of the minimum required independent pad connections, $\sum_n(|P_n|-1)$, that are routed and DRC-clean. Missing outputs remain in both denominators and receive no credit.
We also report the percentages of expected nets affected by opens (Open), zero-clearance physical shorts (PShort), and logical shorts (NShort). PShort includes only different-net wire--wire and via--wire pairs; pad, copper-pour, and via--via contacts are excluded. Missing outputs remain in the error-rate denominators but do not increase the error numerators. The published DRC uses the same clearance checks, so Total equals Clr., while Dist. is zero in the current evaluation.

\begin{figure}[t]
    \centering
    \includegraphics[width=0.9\columnwidth]{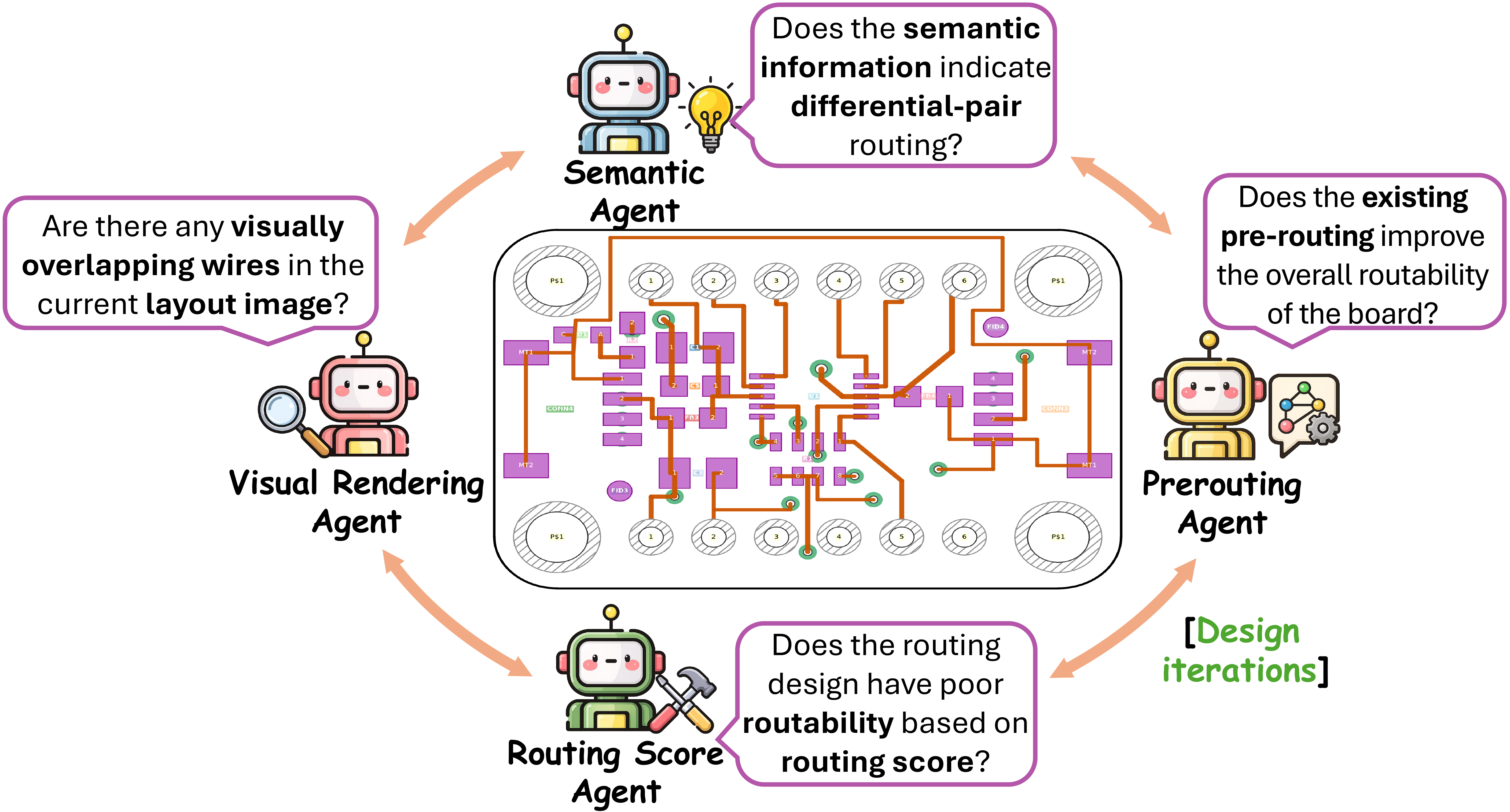}
    \caption{Overview of a multi-modal, multi-agent, multi-round agentic framework for evaluating LMM-based tool usage in PCB routing.}
    \label{fig:react_framework}
\end{figure}

\begin{table*}[!ht]
\centering
\caption{\textbf{Evaluation of LMMs, a classical routing baseline, and the human-engineered reference on PCB routing.} \textit{PR} is the fraction of loadable outputs. \textit{NRR} and \textit{PRR} measure DRC-clean routed nets and required pad-to-pad connections; missing outputs remain in their denominators. \textit{Open}, \textit{PShort}, and \textit{NShort} denote nets affected by opens, physical shorts, and logical shorts. \textit{Clr.} is the mean retained clearance violations per board; \textit{Dist.} is zero because boundary distance is not evaluated. \textit{TWL}, \textit{\#Vias}, \textit{\#Layers}, and \textit{RT} denote wire length, via count, used layers, and runtime. \textit{MGS} measures routing similarity across coarse and fine grids. Higher is better for PR, NRR, PRR, and MGS; lower is better otherwise. Dataset details are provided in Appendix~B. ``-'' denotes unavailable results.
}

\label{tab:oneshot}

\begingroup
\small
\setlength{\tabcolsep}{1mm}
\renewcommand{\arraystretch}{0.6}

\begin{tabular}{l c l c ccccc cc ccc c c}
\toprule

\multirow{2}{*}{Model}
& \multirow{2}{*}{Size}
& \multirow{2}{*}{Expe.}
& \multicolumn{2}{c}{Routability}
& \multicolumn{3}{c}{Connectivity}
& \multicolumn{2}{c}{Design Rules}
& \multicolumn{3}{c}{Routing Quality}
& \multicolumn{2}{c}{Efficiency}
& \multicolumn{1}{c}{Ref.} \\
\cmidrule(lr){4-5}
\cmidrule(lr){6-8}
\cmidrule(lr){9-10}
\cmidrule(lr){11-13}
\cmidrule(lr){14-15}
\cmidrule(lr){16-16}
& & &
\makecell[c]{NRR\\(\%) $\uparrow$}
& \makecell[c]{PRR\\(\%) $\uparrow$}
& \makecell[c]{Open\\(\%) $\downarrow$}
& \makecell[c]{PShort\\(\%) $\downarrow$}
& \makecell[c]{NShort\\(\%) $\downarrow$}
& \makecell[c]{Clr.\\$\downarrow$}
& \makecell[c]{Dist.\\$\downarrow$}
& \makecell[c]{TWL\\$\downarrow$}
& \makecell[c]{\#V.}
& \makecell[c]{\#L.}
& \makecell[c]{PR\\$\uparrow$}
& \makecell[c]{RT\\$\downarrow$}
& \makecell[c]{MGS\\(\%) $\uparrow$} \\

\midrule
\rowcolor{gray!20}
\multicolumn{16}{c}{Commercial LMMs} \\
\midrule

\multirow{5}{*}{\makecell[l]{GPT\\5.5}} & \multirow{5}{*}{-} & Base & \underline{9.13} & \underline{12.26} & \underline{20.09} & 75.37 & \textbf{1.00} & 427.21 & 4.08 & 791.66 & 20.53 & 1.74 & \underline{0.86} & 133.79 & \textbf{15.57} \\
 &  & Net & \underline{9.87} & \underline{12.83} & \underline{20.66} & 75.12 & \textbf{1.14} & 397.07 & 4.24 & 774.93 & 21.02 & 1.79 & 0.86 & 133.02 & \underline{10.36} \\
 &  & Layer & \underline{9.00} & \underline{11.66} & 31.46 & 71.24 & \textbf{0.92} & 362.50 & 4.58 & 781.98 & 22.79 & 4.91 & 0.85 & 130.22 & \textbf{4.53} \\
 &  & Few. & \underline{9.42} & \textbf{13.11} & 19.03 & 80.07 & \underline{1.39} & 342.61 & 4.03 & 685.34 & 10.56 & 1.75 & \underline{0.96} & 117.35 & \underline{5.01} \\
 &  & Pre. & \underline{9.99} & \underline{14.23} & \underline{22.38} & 77.95 & \textbf{1.51} & 349.52 & 4.92 & 735.99 & 27.25 & 1.88 & 0.90 & 127.46 & \textbf{6.41} \\

\midrule

\multirow{5}{*}{\makecell[l]{GPT\\5.4 Mini}} & \multirow{5}{*}{-} & Base & 5.75 & 6.74 & 31.20 & 75.49 & 3.11 & 368.80 & 3.28 & 648.43 & 15.25 & \underline{1.38} & \textbf{0.99} & 69.89 & 11.92 \\
 &  & Net & 5.57 & 6.02 & 38.91 & 68.96 & 4.58 & 291.18 & 3.11 & 552.48 & 8.95 & \underline{1.27} & \textbf{0.99} & 62.52 & 7.99 \\
 &  & Layer & 5.12 & 5.26 & 43.15 & 68.87 & 4.30 & 301.08 & \textbf{2.69} & 567.12 & 11.13 & 2.15 & \textbf{0.99} & 62.61 & 3.82 \\
 &  & Few. & 4.09 & 4.37 & 61.91 & 53.70 & 10.18 & 148.81 & \textbf{1.87} & 323.94 & 3.11 & \underline{1.26} & \textbf{0.99} & 55.87 & 3.64 \\
 &  & Pre. & 5.91 & 6.12 & 40.08 & 66.69 & 4.93 & 269.96 & \textbf{2.73} & 511.56 & 14.44 & \underline{1.31} & \textbf{0.99} & 66.19 & 6.18 \\

\midrule

\multirow{5}{*}{\makecell[l]{Claude\\Opus 4.8}} & \multirow{5}{*}{-} & Base & \textbf{12.60} & \textbf{15.41} & 55.05 & 73.71 & 3.61 & \underline{147.72} & \underline{2.80} & 494.41 & 13.27 & 1.95 & \textbf{0.99} & 42.06 & \underline{14.32} \\
 &  & Net & \textbf{12.38} & \textbf{14.16} & 55.73 & 71.59 & 4.11 & \underline{133.27} & \underline{3.01} & 469.17 & 11.12 & 1.95 & \textbf{0.99} & 39.62 & 9.24 \\
 &  & Layer & \textbf{12.09} & \textbf{13.39} & 59.06 & 69.82 & 3.31 & \underline{121.41} & 2.91 & 468.88 & 15.35 & 4.63 & \textbf{0.99} & 39.99 & 4.33 \\
 &  & Few. & \textbf{10.36} & \underline{11.32} & 44.03 & 62.32 & 4.09 & \textbf{123.83} & \underline{2.45} & 424.89 & 7.72 & 1.89 & 0.95 & 37.76 & 4.46 \\
 &  & Pre. & \textbf{12.54} & \textbf{14.42} & 53.13 & 72.44 & 4.35 & \underline{137.74} & \underline{3.33} & 480.48 & 15.77 & 1.95 & \textbf{0.99} & 41.43 & \underline{6.34} \\

\midrule

\multirow{5}{*}{\makecell[l]{Gemini\\3.1 Pro}} & \multirow{5}{*}{-} & Base & 5.05 & 6.20 & 39.39 & 33.12 & \underline{2.03} & 203.66 & \textbf{2.09} & 448.73 & 14.64 & 1.50 & 0.69 & 170.18 & 13.95 \\
 &  & Net & 4.63 & 6.58 & 36.04 & 35.86 & \underline{3.11} & 248.06 & \textbf{2.66} & 507.48 & 13.72 & 1.55 & 0.70 & 163.17 & \textbf{14.01} \\
 &  & Layer & 4.70 & 6.30 & 36.97 & 36.94 & \underline{2.53} & 251.86 & \underline{2.87} & 533.25 & 14.16 & 2.00 & 0.70 & 164.67 & \textbf{4.53} \\
 &  & Few. & 4.47 & 4.97 & \underline{11.77} & 20.69 & \textbf{0.96} & 242.16 & 2.48 & 668.46 & 18.63 & 1.85 & 0.33 & 274.01 & \textbf{5.16} \\
 &  & Pre. & 3.90 & 6.53 & 25.80 & 42.15 & \underline{3.50} & 344.12 & 3.45 & 664.88 & 20.18 & 1.91 & 0.60 & 188.35 & 6.29 \\

\midrule

\multirow{5}{*}{\makecell[l]{Gemini\\3.5 Flash}} & \multirow{5}{*}{-} & Base & 1.37 & 1.30 & \textbf{16.09} & \underline{25.07} & 9.27 & 414.89 & 4.58 & 427.48 & 42.48 & 1.41 & 0.38 & \underline{21.20} & 9.17 \\
 &  & Net & 1.33 & 1.26 & \textbf{18.16} & \textbf{23.77} & 9.72 & 379.96 & 5.11 & \underline{370.20} & 34.88 & 1.41 & 0.37 & \textbf{22.62} & 8.18 \\
 &  & Layer & 1.12 & 1.03 & \textbf{18.16} & \textbf{23.57} & 9.50 & 358.39 & 5.50 & \underline{361.86} & 31.00 & \underline{1.39} & 0.36 & \textbf{21.67} & 3.91 \\
 &  & Few. & 0.09 & 0.05 & \textbf{6.80} & \textbf{6.37} & 3.49 & 834.67 & 21.59 & 419.05 & \underline{0.03} & 1.71 & 0.13 & \textbf{25.66} & 3.85 \\
 &  & Pre. & 0.77 & 0.67 & \textbf{15.40} & \textbf{18.74} & 7.82 & 433.26 & 6.16 & 370.76 & 37.42 & 1.50 & 0.32 & \underline{27.05} & 4.92 \\

\midrule

\rowcolor{gray!20}
\multicolumn{16}{c}{Open-Source LMMs} \\
\midrule

\multirow{5}{*}{\makecell[l]{Llama4\\Maverick}} & \multirow{5}{*}{400B} & Base & 4.28 & 5.14 & 28.70 & 55.76 & 26.75 & 231.39 & 4.07 & \underline{420.04} & 11.34 & 1.74 & 0.83 & \textbf{20.18} & 13.25 \\
 &  & Net & 3.37 & 3.75 & 33.63 & 55.58 & 34.47 & 234.30 & 3.91 & 382.55 & 11.29 & 1.69 & 0.84 & \underline{25.13} & 7.72 \\
 &  & Layer & 3.58 & 3.70 & 33.30 & 55.89 & 33.74 & 234.65 & 3.51 & 382.18 & 10.75 & 1.59 & 0.85 & \underline{24.99} & 4.12 \\
 &  & Few. & 0.44 & 0.60 & 21.37 & 22.38 & 14.57 & 145.79 & 6.50 & \underline{219.58} & 3.32 & 1.61 & 0.51 & \underline{33.16} & 3.42 \\
 &  & Pre. & 3.03 & 3.25 & 32.70 & 52.61 & 33.16 & 222.23 & 3.73 & \underline{368.43} & 14.35 & 1.75 & 0.82 & \textbf{25.91} & 4.95 \\

\midrule

\multirow{5}{*}{\makecell[l]{Ministral3\\14B 2512}} & \multirow{5}{*}{-} & Base & 3.67 & 4.25 & 31.45 & 52.32 & 21.57 & 252.77 & 4.62 & 432.39 & 9.16 & 1.95 & 0.84 & 84.54 & 12.31 \\
 &  & Net & 3.67 & 3.86 & 30.38 & 54.34 & 23.26 & 272.08 & 4.10 & 438.33 & 8.57 & 1.87 & 0.84 & 86.26 & 8.38 \\
 &  & Layer & 3.68 & 3.95 & \underline{30.70} & 56.57 & 24.70 & 286.56 & 4.09 & 452.39 & 6.25 & 1.58 & 0.85 & 86.75 & 4.13 \\
 &  & Few. & 3.13 & 3.31 & 25.50 & 43.42 & 18.80 & 221.55 & 3.80 & 361.71 & 1.31 & 1.54 & 0.78 & 102.77 & 4.06 \\
 &  & Pre. & 3.63 & 3.87 & 33.36 & 57.63 & 24.33 & 278.88 & 4.67 & 447.21 & 19.13 & 1.91 & 0.86 & 89.05 & 5.61 \\

\midrule

\multirow{5}{*}{\makecell[l]{Qwen\\3.5 9B}} & \multirow{5}{*}{9B} & Base & 1.24 & 1.61 & 62.69 & \textbf{20.45} & 6.48 & \textbf{63.86} & 5.91 & \textbf{131.34} & \textbf{0.09} & \textbf{1.01} & 0.73 & 235.36 & 3.98 \\
 &  & Net & 1.10 & 2.37 & 60.58 & \underline{25.37} & 5.00 & \textbf{74.33} & 3.83 & \textbf{161.32} & \textbf{0.09} & \textbf{1.03} & 0.73 & 183.59 & 3.32 \\
 &  & Layer & 1.28 & 2.36 & 60.81 & \underline{24.48} & 4.93 & \textbf{69.29} & 3.99 & \textbf{158.99} & \textbf{0.08} & \textbf{1.02} & 0.73 & 262.85 & 2.68 \\
 &  & Few. & 1.08 & 1.06 & 27.31 & \underline{12.44} & 3.94 & \underline{124.18} & 3.84 & \textbf{185.32} & \textbf{0.00} & \textbf{1.03} & 0.44 & 758.87 & 3.31 \\
 &  & Pre. & 1.07 & 1.74 & 54.98 & \underline{23.43} & 4.97 & \textbf{85.76} & 4.55 & \textbf{163.74} & \textbf{0.41} & \textbf{1.06} & 0.66 & 266.79 & 2.73 \\

\midrule

\multirow{5}{*}{\makecell[l]{Qwen3\\235B-\\A22B}} & \multirow{5}{*}{\makecell[l]{235B-\\A22B}} & Base & 5.23 & 7.14 & 32.50 & 78.25 & 21.21 & 371.62 & 4.35 & 611.40 & \underline{5.82} & 1.52 & \textbf{0.99} & 114.74 & 13.68 \\
 &  & Net & 4.16 & 5.68 & 33.53 & 64.74 & 17.36 & 327.14 & 4.16 & 540.49 & \underline{5.38} & 1.46 & \underline{0.94} & 212.55 & 8.62 \\
 &  & Layer & 4.40 & 5.74 & 32.41 & 63.75 & 17.93 & 335.54 & 5.58 & 540.17 & \underline{5.13} & 1.44 & \underline{0.93} & 206.66 & \underline{4.39} \\
 &  & Few. & 2.90 & 3.09 & 26.06 & 39.65 & 19.63 & 350.44 & 6.76 & 429.82 & 1.41 & 1.43 & 0.75 & 804.57 & 4.03 \\
 &  & Pre. & 3.64 & 4.48 & 35.36 & 53.93 & 16.55 & 266.62 & 4.35 & 443.68 & \underline{7.07} & 1.50 & \underline{0.91} & 180.73 & 6.27 \\

\midrule

\rowcolor{gray!20}
\multicolumn{16}{c}{Open-Source Routing Baseline} \\

\midrule

PcbRouter & - & Base& 56.20 & 71.37 & 12.93 & 12.97 & 0.85 & 54.85 & 22.62 & 745.20 & 68.15 & 1.96 & 0.993 & 45.64 & 48.32 \\
GPCBRouter & - & Base & 0.69 & - & 99.00 & - & - & - & - & - & - & - & 0.993 & - & - \\

\midrule

\rowcolor{gray!20}
\multicolumn{16}{c}{Human Engineer Reference} \\

\midrule

Human Engineer & - & Ref. & 93.58 & 95.91 & 4.38 & 0.38 & 0.04 & 0.80 & 0.00 & 617.40 & 48.56 & 2.09 & 1.000 & - & 100.00 \\

\bottomrule
\end{tabular}%

\endgroup

\end{table*}

```latex
\begin{table*}[!ht]
\centering
\caption{\textbf{Agentic evaluation of LMMs on PCB routing.} Metric definitions follow Table~\ref{tab:oneshot}. \textit{Avg. Steps} denotes the mean number of agent iterations used to obtain the final routing solution. Missing outputs are retained in the routability denominators. \textit{MGS (\%)} is Multi-Scale Graph Similarity, averaged over outputs with at least one evaluable target net; ``-'' indicates that a result is unavailable. Higher values are better for PR, NRR, PRR, and MGS, whereas lower values are preferred for violation counts, wire length, via count, runtime, and average steps.
}

\label{tab:agentic}

\begingroup
\small
\setlength{\tabcolsep}{1mm}
\renewcommand{\arraystretch}{0.6}

\begin{tabular}{l c l cc ccc ccc ccc ccc c}
\toprule

\multirow{2}{*}{Model}
& \multirow{2}{*}{Size}
& \multirow{2}{*}{Expe.}
& \multicolumn{2}{c}{Routability}
& \multicolumn{3}{c}{Connectivity}
& \multicolumn{3}{c}{Design Rules}
& \multicolumn{3}{c}{Routing Quality}
& \multicolumn{3}{c}{Efficiency}
& \multicolumn{1}{c}{Ref.} \\
\cmidrule(lr){4-5}
\cmidrule(lr){6-8}
\cmidrule(lr){9-11}
\cmidrule(lr){12-14}
\cmidrule(lr){15-17}
\cmidrule(lr){18-18}
& & &
\makecell[c]{NRR\\(\%) $\uparrow$}
& \makecell[c]{PRR\\(\%) $\uparrow$}
& \makecell[c]{Open\\(\%) $\downarrow$}
& \makecell[c]{PShort\\(\%) $\downarrow$}
& \makecell[c]{NShort\\(\%) $\downarrow$}
& \makecell[c]{Clr.\\$\downarrow$}
& \makecell[c]{Dist.\\$\downarrow$}
& \makecell[c]{Polygon\\(\%)}
& \makecell[c]{TWL\\$\downarrow$}
& \makecell[c]{\#V}
& \makecell[c]{\#L}
& \makecell[c]{PR\\$\uparrow$}
& \makecell[c]{RT\\$\downarrow$}
& \makecell[c]{Avg. \\Steps}
& \makecell[c]{MGS\\(\%) $\uparrow$} \\

\midrule

\rowcolor{gray!20}
\multicolumn{18}{c}{Commercial LMMs} \\

\midrule

\multirow{6}{*}{GPT-5.5} & \multirow{6}{*}{-} & Base & \textbf{21.71} & \textbf{24.78} & 57.70 & 39.53 & \underline{0.39} & \textbf{42.87} & 5.19 & 0.00 & 578.95 & 29.11 & 1.97 & \textbf{1.00} & 1060.60 & 16.97 & \underline{18.45} \\
 &  & Vis. & \textbf{20.96} & \textbf{24.46} & 58.40 & 36.99 & \underline{0.39} & \textbf{32.90} & 4.91 & 0.00 & 549.45 & 26.34 & 1.97 & \textbf{1.00} & 1139.14 & 17.09 & \underline{18.89} \\
 &  & Score & \textbf{24.78} & \textbf{20.99} & 59.62 & \underline{10.07} & \underline{0.27} & \textbf{5.58} & 4.01 & 0.00 & 332.51 & 15.50 & 1.96 & \textbf{1.00} & 779.40 & 20.26 & \underline{17.70} \\
 &  & Sem. & \textbf{21.75} & \textbf{25.64} & 56.88 & 38.24 & 0.62 & \textbf{36.82} & 5.09 & 0.00 & 554.48 & 28.10 & 1.96 & \textbf{1.00} & 1037.56 & 16.58 & \underline{11.36} \\
 &  & Pre. & \textbf{22.61} & \textbf{30.34} & 51.10 & 43.11 & \underline{0.42} & \textbf{57.92} & 5.51 & 0.00 & 606.21 & 30.81 & 1.98 & \textbf{1.00} & 988.91 & 15.80 & \underline{30.23} \\
 &  & All & \textbf{27.98} & \textbf{25.60} & 56.54 & 9.12 & \underline{0.26} & \textbf{5.27} & 4.08 & 0.00 & 326.57 & 17.88 & 1.96 & \textbf{1.00} & 737.76 & 20.78 & \underline{18.51} \\

\midrule

\multirow{6}{*}{GPT-5-mini} & \multirow{6}{*}{-} & Base & 9.86 & 7.63 & 56.73 & 56.91 & 8.10 & \underline{88.98} & 2.77 & 0.00 & 289.96 & 11.16 & 1.78 & \textbf{1.00} & 368.76 & 18.38 & 13.74 \\
 &  & Vis. & 9.91 & 7.40 & 57.86 & 53.31 & 7.34 & 78.95 & 1.91 & 0.00 & 247.67 & 8.61 & 1.75 & \textbf{1.00} & 385.25 & 20.00 & 13.96 \\
 &  & Score & 9.85 & 6.35 & 69.29 & 29.86 & 2.22 & 19.92 & 1.24 & 0.00 & 113.28 & 4.56 & 1.75 & \textbf{1.00} & 355.64 & 23.88 & 11.21 \\
 &  & Sem. & 9.08 & 7.15 & 54.25 & 54.24 & 7.76 & 86.92 & 2.21 & 0.00 & 255.54 & 6.86 & 1.69 & \underline{0.99} & 413.85 & 19.42 & 8.33 \\
 &  & Pre. & 9.15 & 10.47 & 36.99 & 43.89 & 5.35 & \underline{74.60} & 2.99 & 0.00 & 309.55 & 15.49 & 1.92 & \underline{0.89} & 360.66 & 17.66 & 25.51 \\
 &  & All & 6.82 & 5.68 & 57.36 & 24.57 & 2.02 & \underline{20.77} & 1.80 & 0.00 & 118.84 & 7.49 & 1.74 & \underline{0.91} & 410.83 & 23.65 & 8.56 \\

\midrule

\multirow{6}{*}{\shortstack{Claude\\Opus 4.8}} & \multirow{6}{*}{-} & Base & \underline{10.65} & \underline{9.27} & 52.63 & 62.76 & 14.59 & 101.89 & 2.43 & 0.00 & 332.91 & 5.64 & 1.71 & \textbf{1.00} & \underline{73.85} & \underline{8.61} & 15.80 \\
 &  & Vis. & \underline{11.18} & \underline{9.50} & 54.48 & 58.21 & 12.64 & 85.02 & 2.83 & 0.00 & 305.63 & 5.64 & 1.76 & \textbf{1.00} & \textbf{75.09} & \textbf{9.13} & 15.50 \\
 &  & Score & \underline{13.44} & \underline{7.49} & 67.22 & 30.18 & 5.08 & 18.17 & 1.35 & 0.00 & 131.14 & 3.27 & 1.74 & \textbf{1.00} & \underline{83.79} & \textbf{9.75} & 13.19 \\
 &  & Sem. & \underline{11.28} & \underline{9.50} & 53.16 & 55.27 & 12.63 & 82.57 & 2.73 & 0.00 & 280.41 & 4.01 & 1.61 & \textbf{1.00} & \underline{74.36} & 8.83 & 9.47 \\
 &  & Pre. & \underline{14.12} & \underline{17.27} & 54.66 & 56.60 & 11.39 & 86.00 & 4.01 & 0.00 & 411.87 & 14.29 & 1.96 & \textbf{1.00} & \textbf{72.88} & \underline{8.08} & 27.20 \\
 &  & All & \underline{14.89} & \underline{13.71} & 63.06 & 32.30 & 5.32 & 23.84 & 2.52 & 0.00 & 200.93 & 10.02 & 1.80 & \textbf{1.00} & \underline{85.43} & \underline{9.77} & 15.20 \\

\midrule

\multirow{6}{*}{\shortstack{Gemini 3.1\\Pro-Preview}} & \multirow{6}{*}{-} & Base & 2.20 & 3.12 & \textbf{10.24} & \underline{10.47} & \textbf{0.17} & 111.00 & 5.24 & 0.00 & 897.01 & 35.19 & 1.95 & 0.19 & 1312.46 & \textbf{6.40} & \textbf{19.40} \\
 &  & Vis. & 1.82 & 2.61 & \textbf{9.40} & \underline{8.69} & \textbf{0.06} & 106.18 & 4.95 & 1.28 & 925.45 & 36.41 & 1.96 & 0.16 & 1692.32 & 12.09 & \textbf{19.18} \\
 &  & Score & 0.44 & 0.41 & \textbf{1.71} & \textbf{1.36} & \textbf{0.00} & 41.61 & 4.11 & 0.00 & 448.71 & 24.61 & 1.94 & 0.04 & 2034.23 & 14.39 & \textbf{22.05} \\
 &  & Sem. & 2.59 & 3.79 & \underline{12.36} & 15.54 & \textbf{0.40} & 131.56 & 4.01 & 4.10 & 864.10 & 32.11 & 1.92 & 0.24 & 1198.73 & \underline{6.05} & \textbf{15.65} \\
 &  & Pre. & 4.08 & 5.28 & \underline{12.08} & 14.12 & \textbf{0.14} & 105.89 & 3.20 & 2.21 & 725.41 & 30.71 & 1.96 & 0.27 & 1131.42 & \textbf{6.42} & \textbf{30.32} \\
 &  & All & 1.17 & 1.01 & \underline{4.65} & \underline{3.14} & \textbf{0.15} & 51.63 & 3.44 & 0.00 & 551.93 & 22.27 & 1.76 & 0.09 & 1876.04 & 15.40 & \textbf{20.83} \\

\midrule

\multirow{6}{*}{\shortstack{Gemini 2.5\\Flash-lite}} & \multirow{6}{*}{-} & Base & 0.27 & 0.20 & 47.17 & \textbf{8.88} & 4.55 & 215.06 & \underline{1.11} & 0.00 & \textbf{98.94} & 2.58 & 1.10 & 0.54 & \textbf{65.48} & 22.51 & 2.96 \\
 &  & Vis. & 0.11 & 0.13 & 51.94 & \textbf{8.39} & 4.62 & 110.52 & \textbf{0.57} & 0.00 & \textbf{61.98} & \textbf{0.27} & 0.86 & 0.62 & \underline{92.98} & 23.34 & 1.48 \\
 &  & Score & 0.23 & 0.17 & 67.86 & 12.45 & 4.97 & 29.39 & 1.05 & 0.00 & \textbf{40.20} & \underline{0.99} & 1.03 & 0.74 & \textbf{64.33} & 24.71 & 1.84 \\
 &  & Sem. & 0.09 & 0.12 & 55.21 & \underline{5.37} & 2.62 & \underline{55.76} & 3.02 & 0.00 & \textbf{40.81} & \textbf{0.38} & 0.40 & 0.66 & \textbf{35.78} & 24.36 & 1.06 \\
 &  & Pre. & 0.47 & 1.10 & 16.92 & \underline{13.37} & 8.14 & 249.48 & 6.33 & 0.00 & \textbf{222.90} & 14.61 & 1.62 & 0.34 & \underline{79.62} & 14.63 & 15.82 \\
 &  & All & 0.46 & 0.71 & 32.99 & 12.37 & 5.07 & 76.97 & 3.76 & 0.00 & 101.73 & 8.81 & 1.38 & 0.49 & \textbf{59.99} & 23.20 & 6.10 \\

\midrule

\rowcolor{gray!20}
\multicolumn{18}{c}{Open-Source LMMs} \\
\midrule

\multirow{6}{*}{\shortstack{LLaMA 4\\Maverick}} & \multirow{6}{*}{400B} & Base & 0.33 & 0.59 & 32.53 & 25.77 & 22.28 & 187.68 & 3.39 & 0.00 & \underline{210.01} & \underline{2.50} & 1.31 & 0.47 & 576.45 & 9.09 & 5.52 \\
 &  & Vis. & 0.26 & 0.47 & 37.71 & 17.48 & 13.20 & \underline{63.61} & \underline{1.15} & 0.00 & \underline{99.09} & \underline{0.88} & 1.21 & 0.48 & 509.38 & \underline{9.68} & 3.92 \\
 &  & Score & 0.20 & 0.29 & 36.78 & 11.20 & 7.54 & \underline{18.05} & \textbf{0.47} & 0.00 & \underline{47.03} & \textbf{0.70} & 1.30 & 0.45 & 531.87 & \underline{9.94} & 2.13 \\
 &  & Sem. & 0.27 & 0.73 & 46.07 & 30.20 & 24.21 & 122.23 & \underline{1.67} & 0.00 & 153.30 & \underline{1.94} & 1.25 & 0.64 & 490.12 & 9.41 & 2.61 \\
 &  & Pre. & 1.40 & 3.75 & 31.08 & 18.74 & 13.20 & 136.93 & \underline{2.11} & 0.00 & \underline{236.45} & \underline{9.42} & 1.81 & 0.45 & 550.79 & 9.20 & 17.01 \\
 &  & All & 1.11 & 2.77 & 49.11 & 13.59 & 7.59 & 21.20 & \underline{0.58} & 0.00 & \textbf{83.54} & \underline{4.48} & 1.48 & 0.62 & 506.25 & 9.97 & 6.97 \\

\midrule

\multirow{6}{*}{\shortstack{Ministral\\14B}} & \multirow{6}{*}{14B} & Base & 0.39 & 0.72 & 70.68 & 43.24 & 23.94 & 226.89 & 4.10 & 0.22 & 230.00 & 4.40 & 1.58 & \underline{0.90} & 399.42 & 9.60 & 3.01 \\
 &  & Vis. & 0.49 & 0.78 & 80.50 & 37.75 & 18.75 & 127.31 & 3.20 & 0.42 & 161.60 & 2.10 & 1.39 & \underline{0.96} & 296.62 & 9.71 & 2.82 \\
 &  & Score & 0.79 & 0.76 & 84.30 & 26.63 & 10.04 & 31.97 & 2.15 & 0.00 & 85.78 & 1.25 & 1.47 & \underline{0.97} & 388.79 & 9.97 & 1.82 \\
 &  & Sem. & 0.27 & 0.50 & 52.18 & 29.91 & 15.70 & 203.78 & 2.20 & 1.06 & 177.15 & 2.03 & 1.30 & 0.75 & 352.02 & 9.67 & 2.08 \\
 &  & Pre. & 0.80 & 2.68 & 58.08 & 43.45 & 24.30 & 230.20 & 4.09 & 0.00 & 258.26 & 14.73 & 1.82 & 0.84 & 356.41 & 9.34 & 5.62 \\
 &  & All & 0.96 & 2.20 & 54.63 & 18.12 & 6.54 & 27.77 & 0.77 & 0.00 & \underline{92.23} & \textbf{4.45} & 1.62 & 0.82 & 345.07 & 10.00 & 5.42 \\

\midrule

\multirow{6}{*}{Qwen3.5-9B} & \multirow{6}{*}{9B} & Base & 1.40 & 0.95 & \underline{20.28} & 15.71 & 6.75 & 129.93 & \textbf{0.94} & 0.76 & 264.21 & \textbf{1.47} & 1.16 & 0.26 & 477.12 & 18.81 & 8.61 \\
 &  & Vis. & 1.85 & 1.00 & \underline{25.54} & 17.57 & 6.82 & 97.20 & 1.35 & 0.00 & 226.80 & 1.06 & 1.19 & 0.33 & 436.80 & 20.53 & 8.89 \\
 &  & Score & 1.39 & 0.76 & \underline{21.80} & 14.80 & 4.78 & 78.95 & \underline{0.48} & 0.00 & 200.14 & 1.17 & 1.35 & 0.27 & 506.80 & 21.29 & 8.44 \\
 &  & Sem. & 0.09 & 0.10 & \textbf{2.66} & \textbf{1.75} & \underline{0.43} & 80.95 & \textbf{0.23} & 0.00 & \underline{144.77} & 3.55 & 1.45 & 0.04 & 241.65 & \textbf{4.86} & 5.84 \\
 &  & Pre. & 0.15 & 0.28 & \textbf{3.56} & \textbf{4.17} & 2.08 & 331.86 & \textbf{0.16} & 0.00 & 371.58 & \textbf{5.27} & 1.66 & 0.09 & 322.00 & 10.80 & 15.54 \\
 &  & All & 0.12 & 0.15 & \textbf{2.56} & \textbf{1.70} & 0.72 & 102.09 & \textbf{0.23} & 0.00 & 178.20 & 5.55 & 1.77 & 0.04 & 149.93 & \textbf{4.95} & 14.42 \\

\midrule

\multirow{6}{*}{\shortstack{Qwen3-235B-\\A22B}} & \multirow{6}{*}{235B} & Base & 0.00 & 0.00 & 0.00 & 0.00 & 0.00 & 0.00 & 0.00 & 0.00 & 0.00 & 0.00 & 0.00 & 0.00 & 63.67 & 1.00 & - \\
 &  & Vis. & - & - & - & - & - & - & - & - & - & - & - & - & - & - & - \\
 &  & Score & - & - & - & - & - & - & - & - & - & - & - & - & - & - & - \\
 &  & Sem. & - & - & - & - & - & - & - & - & - & - & - & - & - & - & - \\
 &  & Pre. & - & - & - & - & - & - & - & - & - & - & - & - & - & - & - \\
 &  & All & - & - & - & - & - & - & - & - & - & - & - & - & - & - & - \\

\bottomrule

\end{tabular}%

\endgroup

\end{table*}
```

For reference-relative evaluation, we introduce \emph{Multi-Scale Graph Similarity} (MGS). Predicted and reference layouts are converted into pad-aware, layer-specific routing graphs at coarse and fine resolutions. The coarse graph captures routing-corridor and spatial-topology agreement, while the fine graph captures local geometry. Greater weight is assigned to the coarse graph, allowing small trace variations while distinguishing different corridors, layers, and via transitions. MGS measures agreement with the expert reference rather than electrical correctness; therefore, a valid alternative route may achieve perfect absolute scores but a lower MGS.
Finally, we report copper-pour usage (Polygon), total wire length (TWL), via count (\#Vias), used routing-layer count (\#Layers), and runtime (RT). \#Layers includes layers used by wires or copper pours but excludes via-only layer extents. Complete definitions, equations, denominators, and graph-construction details are provided in Appendix~B.

\section{Experiment and Findings}

\subsection{Experimental Setups}\label{sec:exp}

\textbf{Study Setup.} The tested LMMs in this section include GPT-5.5~\cite{openai_gpt55}, GPT-5-mini~\cite{openai_gpt5mini}, Gemini-3.1-Pro-Preview~\cite{google_gemini3_pro}, Gemini-2.5-Flash-Lite~\cite{google_gemini_25_flash}, Claude-Opus-4.8~\cite{anthropic_claude_opus_47}, LLaMA-4-Maverick~\cite{meta_llama4}, Mistral-Medium-3.5 (128B)~\cite{mistral_medium35}, Qwen3.5-9B~\cite{qwen35_9b}, and Qwen3.6-35B-A3B~\cite{qwen3-35b-a3b}. We evaluate these models alongside two open-source PCB routing baselines. We conduct two studies: \textbf{(i)} \textbf{Zero-shot Evaluation}, where all LMMs are prompted once without intermediate interaction to assess end-to-end PCB component routing. All settings include the board boundary, netlist, and component dimensions, while we further study the impact of additional information, including schematic images, schematic-aware netlist graphs, global schematic positions, and few-shot examples; and \textbf{(ii)} \textbf{Agentic Evaluation}, where LMMs are equipped with layout visualization, overlap checking, and routability verification tools to iteratively refine PCB routing under geometric and electrical constraints.


\textbf{Implementation of the Prior PCB Routing Baseline.}
To comprehensively evaluate the difficulty and quality of our benchmark, we implement PCB router \cite{openroad_pcbr} and GPCB \cite{10734385} as baselines. We apply the router to all benchmark instances and evaluate its routing performance using the same design constraints and evaluation metrics as our method.

\textbf{Implementation of LMMs Agentic Framework.} 
We design a multi-modal, multi-agent, and multi-round agentic evaluation framework, as shown in Fig.\ref{fig:react_framework}, where 
LMMs are provided with the board boundary, netlist, component 
dimensions, and schematic diagram, and equipped with three 
tools: layout visualization for rendering the current 
routing state, overlap checking for reporting constraint 
violations, and routability verification for estimating 
routing conflicts. At each step, the LMM observes the 
rendered layout and iteratively refines component positions, 
orientations, and layer assignments until all constraints 
are satisfied or the maximum number of steps is reached.

\subsection{Main Results and Findings}
\textbf{Zero-shot and agentic results} are reported in
Tables~\ref{tab:oneshot} and~\ref{tab:agentic}. LMMs remain far below the human
reference: the best zero-shot NRR/PRR is \(12.60\%/15.41\%\), versus
\(93.58\%/95.91\%\) for humans, while agentic refinement reaches only
\(27.98\%/30.34\%\).

\textbf{$\bullet$ Routing Legality.} 
Base LMM outputs average \(69.77\)--\(431.29\) DRC violations and
\(20.45\%\)--\(78.25\%\) PShort, versus \(0.80\) and \(0.38\%\) for humans.
GPT-5.5 All Tools reduces Total from \(48.06\) to \(9.35\) and PShort from
\(39.53\%\) to \(9.12\%\), demonstrating strong local repair but incomplete
legality.

\textbf{$\bullet$ Routability.}
PR measures loadability, not routing success: GPT-5-mini and Qwen3-235B-A22B
reach PR \(=0.989/0.991\) but only \(5.75\%/5.23\%\) NRR. Even the best
zero-shot LMM result trails PcbRouter (\(56.20\%\) NRR and \(71.37\%\) PRR),
showing that valid files often lack complete DRC-clean connections.

\textbf{$\bullet$ Electrical Functionality.} 
Claude Opus 4.8 Base records \(55.05\%\) Open, \(73.71\%\) PShort, and
\(3.61\%\) NShort, versus \(4.38\%/0.38\%/0.04\%\) for humans. Low error rates
can be misleading: Gemini 2.5 Flash-lite Few-shot achieves \(6.80\%\) Open and
\(6.37\%\) PShort but only PR \(=0.132\) and NRR \(=0.09\%\), so connectivity
metrics must be interpreted jointly.

\textbf{$\bullet$ Ablation Study Analysis.} 
Netclass changes average NRR/PRR by only \(-0.25/-0.39\) points while increasing
Open/NShort by \(1.16/0.86\) points. Multi-layer lowers MGS for every model
(GPT-5.5: \(15.57\%\!\rightarrow\!4.53\%\)); Few-shot can reduce violations by
routing less (LLaMA 4: Total \(235.45\!\rightarrow\!152.29\), but NRR
\(4.28\%\!\rightarrow\!0.44\%\)).

\textbf{$\bullet$ Agentic Spatial \& Routing Reasoning.} 
Visual feedback lowers Total for every model, and Score reduces its cross-model
mean from \(141.18\) to \(32.31\) (\(-77.1\%\)). Preroute raises MGS for all
eight models by \(9.97\) points on average (GPT-5.5:
\(18.45\%\!\rightarrow\!30.23\%\)), indicating that tools improve local
legality and routing-corridor selection more than global connectivity.

\textbf{$\bullet$ Reasoning Steps.} 
More steps do not guarantee improvement: GPT-5.5 All Tools increases steps
from \(16.97\) to \(20.78\) and NRR from \(21.71\%\) to \(27.98\%\), whereas
GPT-5-mini increases steps from \(18.38\) to \(23.65\) but reduces NRR from
\(9.86\%\) to \(6.82\%\). Feedback quality matters more than iteration count.
\section{Conclusion}
We introduce \textit{\name}, the first large-scale benchmark for evaluating LMMs on PCB routing reasoning under realistic geometric, topological, electrical, and design-rule constraints. The benchmark contains 1,681 industrial-grade, schematic-coupled PCB designs with engineer-created routable placements, netlists, stackup information, routing constraints, and fine-grained reference routings. We evaluate state-of-the-art LMMs in both end-to-end and tool-augmented settings, together with open-source and commercial routing systems as baselines. Our results show that current LMMs remain substantially limited in multi-net path planning, design-rule compliance, layer and via assignment, congestion handling, and preservation of schematic-defined connectivity. Their generated routes also deviate markedly from expert reference designs in routing topology, trace geometry, layer usage, and manufacturability. We hope that \textit{\name}, together with its evaluation framework and tool interfaces, will support future research on reliable and constraint-aware routing agents for electronic design automation.

\bibliography{aaai2027}


\end{document}